\documentclass[runningheads]{llncs}
\pdfoutput=1

\usepackage[utf8]{inputenc}
\usepackage[T1]{fontenc}
\usepackage{textcomp}

\usepackage{graphicx}
\usepackage{amsmath,amssymb}
\usepackage{color}

\usepackage{microtype}
\usepackage{nicefrac}
\usepackage{amsfonts}
\usepackage{booktabs}
\usepackage{multirow}
\usepackage[table]{xcolor}
\usepackage{tikz}
\usetikzlibrary{calc,spy}
\usepackage{tabularx}
\usepackage{array}

\usepackage[numbers,sort]{natbib}
\usepackage[english]{babel}     %
\usepackage[strings]{underscore} %
\makeatletter\renewcommand{\@biblabel}[1]{#1.}\makeatother 

\usepackage[rightcaption]{sidecap}
\usepackage{wrapfig}

\usepackage[ruled]{algorithm2e}

\usepackage{enumitem}
\setlist{noitemsep,topsep=1pt,parsep=1pt,partopsep=1pt}

\usepackage{url}            %
\usepackage[pagebackref=false,breaklinks,colorlinks,bookmarks=false]{hyperref}

\usepackage{capt-of}
\usepackage[capitalise,noabbrev,nameinlink]{cleveref}
\creflabelformat{equation}{#2#1#3}

\renewcommand{\orcidID}[1]{\href{https://orcid.org/#1}{~\includegraphics[width=9pt]{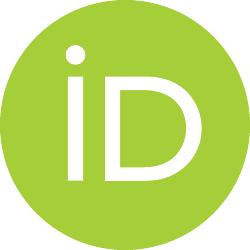}}}

\makeatletter
\renewcommand\paragraph{\@startsection{paragraph}{4}{\z@}%
	{1.75ex \@plus.5ex \@minus.2ex}%
	{-1em}%
	{\normalfont\normalsize\bfseries}}
\makeatother

\AtBeginDocument{%
\setlength\abovedisplayskip{5pt plus 1pt minus 2.5pt} %
\setlength\belowdisplayskip{5pt plus 1pt minus 2.5pt} %
\setlength\abovedisplayshortskip{0pt plus 3pt} %
\setlength\belowdisplayshortskip{3pt plus 1.5pt minus 1.5pt} %
}

\usepackage{mathtools}
\usepackage{xspace}

\newcommand{\abs}[1]{\left\lvert#1\right\rvert}
\newcommand{\norm}[1]{\left\lVert#1\right\rVert}

\newcommand{\ignore}[1]{}

\makeatletter
\DeclareRobustCommand\onedot{\futurelet\@let@token\@onedot}
\def\@onedot{\ifx\@let@token.\else.\null\fi\xspace}

\makeatother

\newcommand{\IGNORE}[1]{}
\newcommand{\degree}{°\xspace}

\DeclareMathOperator{\render}{Render}

\newcommand{\triup}{\textcolor{gray}{$\blacktriangle$}}
\newcommand{\tridown}{\textcolor{gray}{$\blacktriangledown$}}

\begin{document}
\pagestyle{headings}
\mainmatter
\def\ECCVSubNumber{1737}

\title{MatryODShka: Real-time 6DoF Video\\View Synthesis using Multi-Sphere Images}

\titlerunning{MatryODShka: Real-time 6DoF Video using MSIs}
\author{Benjamin~Attal\inst{1,2}\orcidID{0000-0002-0132-5232} \and
	Selena~Ling\inst{1}\orcidID{0000-0001-6458-4488} \and
	Aaron~Gokaslan\inst{1}\orcidID{0000-0002-3575-2961} \and
	Christian~Richardt\inst{3}\orcidID{0000-0001-6716-9845} \and 
	James~Tompkin\inst{1}\orcidID{0000-0003-2218-2899}}
\authorrunning{B.~Attal, S.~Ling, A.~Gokaslan, C.~Richardt, and J.~Tompkin}
\institute{\noindent$^1$Brown~University,~USA~~
	$^2$Carnegie~Mellon~University,~USA~~
	$^3$University~of~Bath,~UK
}

\maketitle
\begin{abstract}
\vspace{-0.25cm}
We introduce a method to convert stereo 360\degree (omnidirectional stereo) imagery into a layered, multi-sphere image representation for six degree-of-freedom (6DoF) rendering.
Stereo 360\degree imagery can be captured from multi-camera systems for virtual reality (VR), but lacks motion parallax and correct-in-all-directions disparity cues.
Together, these can quickly lead to VR sickness when viewing content.
One solution is to try and generate a format suitable for 6DoF rendering, such as by estimating depth.
However, this raises questions as to how to handle disoccluded regions in dynamic scenes.
Our approach is to simultaneously learn depth and disocclusions via a multi-sphere image representation, which can be rendered with correct 6DoF disparity and motion parallax in VR.
This significantly improves comfort for the viewer, and can be inferred and rendered in real time on modern GPU hardware.
Together, these move towards making VR video a more comfortable immersive medium.
\looseness-1
\end{abstract}
\section{Introduction}
\vspace{-0.15cm}

\begin{figure}[t]
    \centering
    \includegraphics[width=1.0\linewidth]{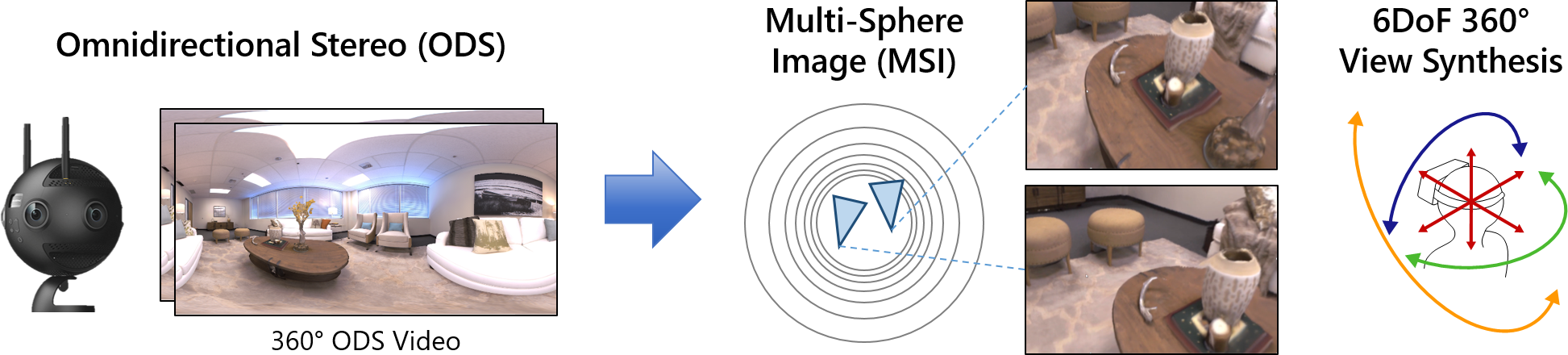}
    \vspace{-6mm}
	\caption{%
			Our approach takes 360\degree omnidirectional stereo video as input and predicts multi-sphere images that enable six degree-of-freedom 360\degree view synthesis in real time.
			This produces a more comfortable and immersive VR video viewing experience.
	}
    \label{fig:bigteaser}
\end{figure}

360\degree imagery is a valuable tool for virtual reality (VR) as the viewer is immersed in a captured real-world environment.
Stereo 360\degree imagery aims to increase this immersion by providing binocular disparity as a depth cue in all directions, and video provides depiction for dynamic scenes.
This imagery is usually captured with wide-angle multi-camera systems, arranged in a circle \cite{AnderGBKSHAS2016,SchroBS2018}, with state-of-the-art systems providing high-resolution and high-quality stitched imagery.
Stereo 360\degree cameras are decreasing in cost ($\approx$\$5k), which will increase their deployment across many industries.
However, there are problems with this format. Motion parallax is missing from stereo 360\degree imagery, which can cause viewing discomfort.
Further, stereo 360\degree formats have disparity problems: side-by-side equirectangular projection (ERP) formats have incorrect disparity everywhere but in one direction \cite{LaiXLL2019}, and omnidirectional stereo (ODS) formats \cite{IshigYT1992,PelegBP2001} have diminished disparity as the view approaches the zenith or nadir \cite{AnderGBKSHAS2016}.
In short, long-term viewing is difficult as vestibulo-ocular comfort is low \cite{ThattG2018}, which can cause VR sickness \cite{PadmaRSNW2018}.

Our goal is to provide six-degree-of-freedom (6DoF) video with accurate motion parallax and disparity cues for a reasonably-sized headbox.
Large-baseline camera systems can \emph{interpolate} views to provide this via optical flow or depth-based reprojection, but usually the desired human motion is too large to build practical camera systems that operate in this way.
Thus, we must \emph{extrapolate} views beyond the camera system's baseline.
This requires estimating content for unseen regions via hallucination or inpainting.
Further, for video, we want this view synthesis to happen quickly, preferably in real time when applied to a stereo 360\degree camera feed, so as to avoid preprocessing and allow live applications.

Our approach is to simultaneously estimate depth and inpaint the holes by using a learning-based approach on a layer-based scene representation.
Inspired by recent work on stereo magnification~\cite{ZhouTFFS2018,SriniTBRNS2019} and light field fusion~\cite{MildeSOKRNK2019}, we learn to decompose a scene into multi-sphere images (MSI), each with RGB and alpha (RGBA) values.
This is created by a network architecture which supports stereo 360\degree input in the omnidirectional stereo format, uses spherically-aware convolutions and losses, and maintains temporal consistency for video without additional network parameters via spherical single-image transform-inverse regularization~\cite{EilerMU2019}.
We demonstrate quantitatively and qualitatively
that these contributions increase reconstruction quality both spatially and temporally against existing view expansion methods, and that our approach can be applied and rendered in real time to 4K videos on modern GPUs.
Our contributions are:

\begin{itemize}%
    \item A multi-sphere image scene representation for omnidirectional view synthesis.
    
    \item A method to recover the MSI representation from ODS imagery via a learning-based soft spherical 3D reconstruction method. This uses an architecture and losses for spherical images, including spherical temporal consistency.
    
    \item A real-time inference and VR rendering engine for MSI from ODS input.
\end{itemize}
These are complemented by an open-source system, with mono (ERP) and stereo 360\degree (ODS) renderers to generate synthetic training data~\cite{SavvaKMZWJSLKMPB2019,StrauWMCWGEMRVCYBYPYZLCBGMPSBSNGLN2019,Googl2015}, TensorFlow models, real-time TensorFlow and TensorRT inference within Unity that outputs to GPU textures, and a real-time multi-sphere video renderer in Unity.
Please see our project webpage at \href{http://visual.cs.brown.edu/matryodshka}{visual.cs.brown.edu/matryodshka}.

\section{Related Work}
\vspace{-0.1cm}

\textbf{360° video stitching}
builds on seminal work in panorama image stitching \cite{BrownL2007,Szeli2006}, which automatically aligns and blends multiple photos of a scene into a single, wide field-of-view panorama.
Subsequent work on stitching 360° videos \cite{LeeKKKN2016,PerazSZKWWG2015} addresses temporally coherent stitching from multi-view video input, as commonly used in commercial 360° videos.
However, monocular 360° videos only provide views for a single center of projection, and hence no depth perception.
Omnidirectional stereo (ODS) is a circular projection \cite{IshigYT1992,PelegBP2001,Richa2020} that improves depth perception using the disparity between the left- and right-eye panoramic views.
ODS has become popular for stereo 360° \cite{AnderGBKSHAS2016,RichaPZS2013,SchroBS2018} as it is a good fit for existing processing, compression and transmission pipelines.

\textbf{360° depth estimation}
aims to recover dense 360° depth maps, which can be used for rendering novel views using a mesh-based renderer.
Assuming a moving camera in a static environment, structure-from-motion and multi-view stereo can be used \cite{ImHRJCK2016,HuangCCJ2017}.
However, the made assumptions are actually violated by most usage scenarios, like stationary cameras or dynamic environments.
Learning-based depth estimation approaches have the potential to overcome these limitations by using single-image input to predict 360° depth maps \cite{LaiXLL2019,WangHCLYSCS2018,ZioulKZD2018,ZioulKZAD2019}.
Nevertheless, view synthesis from RGBD (RGB+Depth) data is fundamentally limited by 3D reconstruction accuracy, and one cannot look behind occlusions \cite{HuangCCJ2017,LaiXLL2019}.

\textbf{360° view synthesis}
creates new panoramic viewpoints from different input \cite{RichaTHHSW2017}.
For example, ODS video can be created from three fisheye cameras \cite{ChapdR2013}, two 360° cameras mounted side by side \cite{MatzeCEKS2017}, or two rotating line cameras \cite{KonraDMW2017}.
However, ODS provides only binocular disparity and no motion parallax.
Novel-view synthesis with motion parallax can be achieved using depth-augmented ODS \cite{ThattBLG2016}, flow-based blending \cite{LuoXRY2018,BerteCR2019}, or a layered scene representation \cite{SerraKCDGHM2019}.
However, these approaches do not support up/down motion.
Parra Pozo et al.'s spherical video camera rig enables high-quality 6DoF view synthesis \cite{ParraTFHMSSC2019}, but not in real time. Serrano et al. similarly propose an offline, optimization-based method for high-quality 6DoF video generation from RGBD equirectangular video input \cite{SerraKCDGHM2019}.
We create a fast learning-based view synthesis method that is applicable to ‘in the wild’ ODS videos or streams, e.g., including all YouTube ODS videos.

\textbf{Perspective view synthesis}
has made leaps in visual quality using soft 3D reconstructions \cite{PenneZ2017,ChoiGTKK2019} and multi-plane images \cite{ZhouTFFS2018,SriniTBRNS2019,FlynnBDDFOST2019,MildeSOKRNK2019}.
MPIs are stacks of semi-transparent layers representing scene appearance without explicit geometry, and can easily be reprojected into novel views.
Learning-based approaches optimize MPIs from stereo images \cite{SriniTBRNS2019,ZhouTFFS2018}, or 4–5 input images \cite{FlynnBDDFOST2019,MildeSOKRNK2019}.
Most approaches optimize RGBA colors per layer; the alpha channel allows for `soft' reconstructions by blending the layers for perspectives from different input views \cite{PenneZ2017}.
We extend these ideas to multi-sphere images, a layered spherical scene representation that enables omnidirectional 6DoF view synthesis.

\textbf{CNNs on spheres}
need to be adapted to correctly handle the unique distortions of 360\degree images. %
Su and Grauman work directly on equirectangular images using wider kernels near the poles \cite{SuG2017a}, but there is no information shared between kernels.
360\degree images can also be projected into a cubemap, processing all sides as perspective images, and recombined \cite{ChengCDWLS2018}.
More principled are distortion-aware convolutions \cite{TatenNT2018,CoorsCG2018,SuG2019}, which also allow transfer of perspectively trained models to equirectangular images.
Full rotation-equi\-variance can be achieved with spherical convolutions \cite{CohenGKW2018,EstevAMD2018}, but this is not necessarily desirable as videos can exploit the fixed gravity vector.
Recent work generalizes cubemaps to icosahedra, and the resulting 20 triangles are unwrapped into five rectangles with shared convolution kernels \cite{LeeJYJY2019,ZhangLSC2019}.
These approaches add expense at inference time, or trade model capacity for spherical awareness; neither of which is desirable.

\section{Method}
\label{sec:method}

\begin{figure*}[t]
    \centering
    \includegraphics[width=\linewidth]{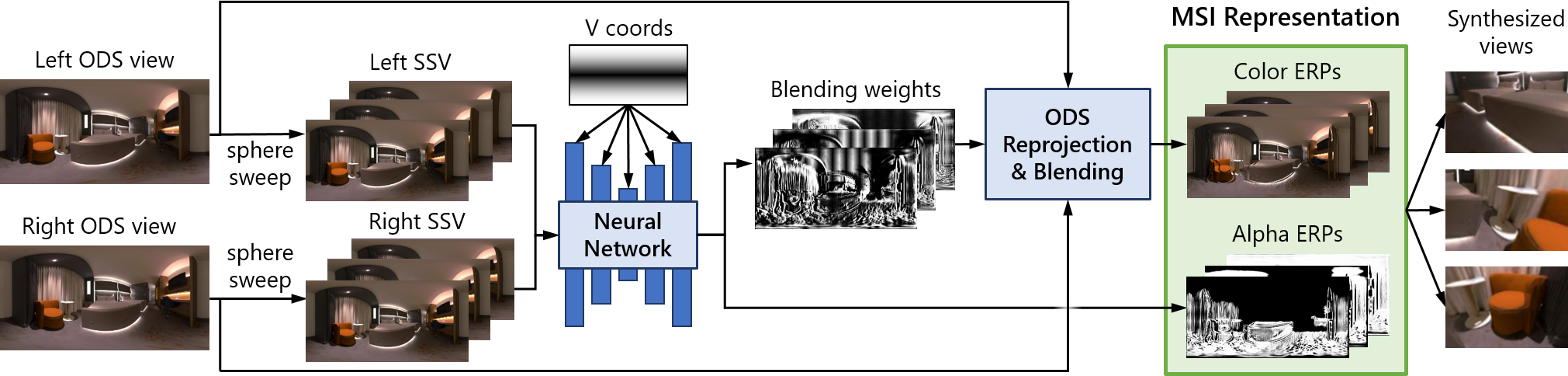}
    \caption{%
		Given an ODS image, we generate left/right sphere sweep volumes (SSV).
		These are input to a fully-convolutional neural network, with V-coordinate convolution, that predicts blending weights and alpha ERPs per multi-sphere image (MSI) layer.
        The final high-res MSI enables real-time 6DoF view synthesis.
        (Figure inspired by Zhou et al.~\cite{ZhouTFFS2018}.)
        \looseness-1
    }
    \label{fig:overview}
\end{figure*}

Our goal is to enable real-time 6DoF view synthesis in the vicinity of an input stereo 360\degree video (\cref{fig:overview}).
We begin with omnidirectional stereo (ODS) imagery, which has an image for each eye given a position in the world \cite{IshigYT1992,PelegBP2001,Richa2020}.
Given a database of synthetic ODS image pairs \cite{Googl2015}, our method trains a network to generate a multi-sphere image (MSI) representation of the scene.
Then, in VR, we infer an MSI for each ODS video frame of an input video, and render it from novel headset viewpoints for the left and right eyes of the user.

\label{sec:projection}
{\setlength\intextsep{0pt}%
\begin{wrapfigure}[6]{r}{0.3\textwidth}
	\includegraphics[width=\linewidth]{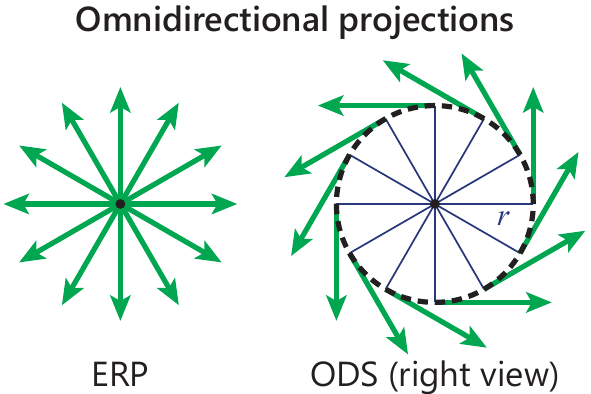}
\end{wrapfigure}

\paragraph{Equirectangular projection (ERP):}
Pixel coordinates in equirectangular images directly map to directions on the unit sphere.
A pixel's $x$-coordinate maps to the azimuth angle $\theta = \pi\times(2x - 1) \in [-\pi, \pi]$, and its $y$-coordinate to the elevation angle $\varphi = \pi\times(\frac{1}{2} - y) \in [-\frac{\pi}{2}, \frac{\pi}{2}]$.
A point $\mathbf{p} = (p_x, p_y, p_z)$ projects to:
\begin{align}
    \theta_\text{ERP}  = - \arctan \!\left( p_z / p_x \right) \quad \text{and} \quad
    \varphi_\text{ERP} = \arctan \!\left( p_y / \sqrt{p_x^2 + p_z^2} \right) \text{,} \label{eq:ERP}
\end{align}
with $x$-forward, $z$-left, and $y$-down \cite{AnderGBKSHAS2016}.
A major disadvantage of the ERP format for 360\degree stereo imagery is that disparity is zero along the camera baseline, and so depth cannot be determined for pixels that lie along the baseline~\cite{MatzeCEKS2017}.

{\setlength\intextsep{0pt}%
\begin{wrapfigure}[6]{r}{0.4\textwidth}
\includegraphics[width=\linewidth]{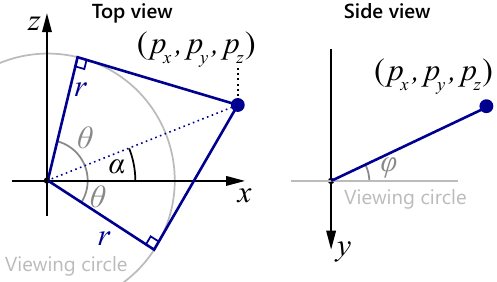}
\end{wrapfigure}

\paragraph{ODS projection:}
This is defined by the \emph{viewing circle} to which all camera rays are tangent \cite{IshigYT1992,PelegBP2001}.
Without loss of generality, this circle lies in the $x$-$z$-plane, is centered at the origin and has radius $r = \text{IPD}/2 = 31.5$\,mm \cite{AnderGBKSHAS2016,Dodgs2004}.
A 3D point $\mathbf{p} = (p_x, p_y, p_z)$ projects into the left and right ODS images at:
\begin{align}
    \theta_\text{ODS}^\text{L/R} &= \pm \arcsin\!\left( r / \sqrt{p_x^2 + p_z^2} \right) - \arctan\!\left( p_z / p_x \right) \text{,}  \label{eq:ODS-theta} \\
    \varphi_\text{ODS} &= \arctan\!\left( p_y / \sqrt{p_x^2 + p_z^2 - r^2} \right) \text{.} \label{eq:ODS-phi}
\end{align}
Unlike ERP, ODS encodes disparity in all azimuth directions,
and therefore is richer input for view-synthesis tasks. %
}%

\subsection{Multi-Sphere Image Representation}
\label{sec:msis}

The core representation for our scene inference and view-synthesis pipeline is the multi-sphere image (MSI)---the spherical equivalent of multi-plane images \cite{ZhouTFFS2018}.
MSIs represent a scene as concentric spheres with color and transparency (RGBA) at each point on a sphere.
One advantage of MSIs is that they allow for fast and dense rendering of novel views using traditional graphics pipelines (e.g., in Unity). %
Unlike multi-plane images, multi-sphere images are omnidirectional and thus enable view synthesis for any camera orientation and position within the innermost sphere.
During training, inference, and rendering, we store MSIs as a sequence of ERPs parametrized by spherical coordinates $\{(\theta_i, \varphi_i)\}$ and their sphere radii $\{r_i\}$.
In practice, we use 32 layers as a trade-off between inference speed and quality, with the smallest radius being 1\,m, and the largest being 100\,m.
We generate the radii for in-between layers by linearly interpolating reciprocal depths.
This creates more layers for nearby scene geometry (half the layers closer than 2\,m).
This sampling pattern is also desirable as it separates layers linearly in disparity when projected into a translated view close to the center of projection. %
See concurrent work by Broxton et al. \cite{BroxtFOEHDDBWD2020} for a theoretical analysis.

\paragraph{Differentiable Rendering from MSIs:}
\label{sec:differentiablemsis}

Novel views can easily be rendered from MSIs for a range of projections, including perspective, ERP and ODS.
We use this to render \textit{target} views for supervision during training.
Each pixel in the projection defines a ray, whose color is given by repeated over-compositing \cite{PorteD1984} on the MSI, similar to MPIs \cite{FlynnBDDFOST2019}.
Each such ray is intersected with the concentric spheres of the MSI, producing intersection points $\{\mathbf{p}_i\}_{i=1}^{N}$ with spherical layers 1 to $N$, from near to far.
Next, each intersection point $\mathbf{p}_i$ is converted from Cartesian to spherical coordinates $(\theta_i, \varphi_i)$, so we can sample the RGB colors $\{\mathbf{c}_i\}_{i = 1}^{N}$ and alphas (opacities) $\{\alpha_i\}_{i = 1}^{N}$ corresponding to these points from MSI layer $i$.
We use bilinear interpolation for sub-pixel precision.
The final color is:
\begin{align}
    \mathbf{c} &= \sum_{i = 1}^{N} \mathbf{c}_i \cdot \underbracket[0.5pt]{\alpha_i \cdot \prod_{j = 1}^{i - 1} (1 - \alpha_j)}_{\text{Net opacity of layer } i} \text{.} \label{eq:overcomp}
\end{align}

\vspace{-0.5cm}
\subsection{Model Architecture}
\label{sec:modelarchitecture}

The goal of our deep model is to infer an RGBA MSI from a pair of ODS images (\cref{fig:overview}).
We use a U-Net-style architecture \cite{RonneFB2015,ZhouTFFS2018} to perform MSI inference, with specific adjustments for the spherical domain.
Our approach of inferring MSI alphas via a new ODS reprojection component conceptually corresponds to a soft 3D reconstruction \cite{PenneZ2017}, and is similar in idea to depth probability volumes~\cite{ChoiGTKK2019}.

Beyond alpha, we structure our network to additionally learn a blending weight between the left and right ODS views for each layer of the MSI, to be used within our reprojection method.
This allows the network to blend between views as appropriate, and to fill holes with content from at least one input view.
This lets us overcome occlusions in one view.
In principle, this also handles specular highlights, reflections, and transparent content that does not correspond between views and so does not have a natural disparity.
This is also useful during training or when inference is imperfect: any ghosting from combining left/right ODS views is minimized when the inferred alphas are opaque at or beyond the correct scene depth, and when the alphas are transparent in front of the correct scene depth.

\paragraph{ODS reprojection:}
We first preprocess a left/right ODS pair $I_L$ and $I_R$
into a pair of \emph{sphere sweep volumes} \cite{ImHRJCK2016}: these are defined as a set of ERP images, each corresponding to the reprojection of concentric spheres with radii $\{r_i\}_{i=1}^{N}$ into one of the ODS images.
We generate each ERP in the sphere sweep as follows: 
\begin{enumerate}
    \item Back-project ERP pixels $(\theta_i, \varphi_i)$ to points $\mathbf{p}_i$ on the sphere of radius $r_i$.
    \item Project these points into the image $I_L$ or $I_R$ (\cref{eq:ODS-theta,eq:ODS-phi}).
    \item Look up pixel colors from $I_L$ or $I_R$ (bilinearly interpolated). %
\end{enumerate}

\paragraph{Blending:}
For each MSI layer, our network predicts alpha values and left/right ODS blending weights.
Specifically, let $\mathcal{S}^\text{L/R} = \{S^\text{L/R}_i\}_{i = 1}^N$ be the sphere sweep for the left/right ODS image, and let $\{\beta_i\}_{i = 1}^{N}$ be the per-layer blending weights.
Then, the colors $\{\mathbf{c}_i\}$ for MSI layer $i$ are calculated via element-wise multiplication ($\odot$):
\begin{align}
    \mathbf{c}_i = \beta_i \odot S^\text{L}_i + (1 - \beta_i) \odot S^\text{R}_i \text{.} \label{eq:odsblend}
\end{align}

\paragraph{Angle-aware kernels:}
To create a training and inference approach which is efficient and implicitly aware of the angular distortions within ERP and ODS images, we provide the network with information about each pixel's relative location in the spherical projection using coordinate convolution \cite{LiuLMSFSY2018}.
Existing approaches provide two additional channels, $U$ and $V$, to each convolutional layer within the network, with each containing the normalized azimuth and elevation at each pixel position \cite{ZioulKZAD2019}.
However, the shape of the image distortion under ERP/ODS projection is independent of azimuth, and is symmetric in elevation around the equator.
Thus we only use $V(x, y) = \abs{\sin(\varphi(y))}$.

\subsection{Training Losses}

During training, we learn to assign alpha values and blending weights to each MSI layer by penalizing the $L_2$ re-rendering error between a predicted target view $\hat{I} = \render(\text{MSI},P)$ for pose $P$ and the ground-truth image $I_\text{GT}$ at $P$:
\begin{align}
\mathcal{L}_{\text{L2}} &= \sum_{x,y} \norm{ \hat{I}(x,y) - I_\text{GT}(x,y) }_2^2.
\end{align}

\paragraph{Spherical weighting:}
Applying an $L_2$ loss directly on equirectangular images puts disproportionate weight on regions near the poles as the projection does not conserve area.
Instead, we use a spherical weighting scheme which weights pixels by their area on the sphere's surface.
We generate a map of area weights $A$ by projecting corner coordinates at pixel $(x,y)$ into spherical coordinates $(\theta_0, \theta_1)$ and $(\varphi_0, \varphi_1)$, and then computing their subtended area on the unit sphere ($r=1$):
\begin{align}
    A(x,y) &= r^2(\theta_1 - \theta_0)(\cos(\varphi_1) - \cos(\varphi_0)) \text{.}
    \label{eq:sphareaweight}
\end{align}
Given a target image $I_{GT}$ and the predicted image $\hat{I} = \render(\text{MSI},P)$ from the MSI at pose P, we then %
apply the spherical area weighting $A$ to the $L_2$ loss:
\begin{align}
    \mathcal{L}_\text{ERP-L2} &= \sum_{x,y} \norm{A(x,y) \cdot \left(\hat{I}(x,y) - I_{GT}(x,y) \right) }_2^2 \text{.}
\end{align}

\paragraph{Transform-inverse MSI regularization:}
Per-frame processing can lead to undesirable flickering in videos.
We improve the temporal consistency of our model with a spherical 3D procedure derived from the 2D image transform-inverse regularization approach \cite{EilerMU2019}.
The motivating idea is that predicting an image under a small transformation of the target view, and then transforming it back to the original target view, should only incur a small difference in appearance.
Penalizing this difference then leads to smoothness over time, as small transformations mimic the minor frame-to-frame differences in a video.
\emph{Single-frame} temporal consistency methods are more efficient to train and infer than two-frame optical-flow-based methods~\cite{BonneTSSPP2015} or recurrent network methods~\cite{LaiHWSYY2018}.

We develop a new approach to apply this to MPIs and MSIs. The input to our network is 3D rather than 2D, via two sphere sweep volumes, $\mathcal{S}^\text{L}$ and $\mathcal{S}^\text{R}$.
Likewise, our output is a 3D representation of a scene. As such, let us consider a 3D rigid body transformation on the inputs and outputs $T = [R \mid t]$, and let $f$ be the function to infer our MSI representation. We would like to compute the loss:
\begin{align}
    \mathcal{L}_{\text{TI}} = \norm{ f(T(\mathcal{S}^\text{L}), T(\mathcal{S}^\text{R})) - T(f(\mathcal{S}^\text{L}, \mathcal{S}^\text{R})) }_2 \text{.}
\end{align}

\setlength\intextsep{0pt}%
\begin{wrapfigure}[10]{r}{0.3\textwidth}
	\includegraphics[width=\linewidth]{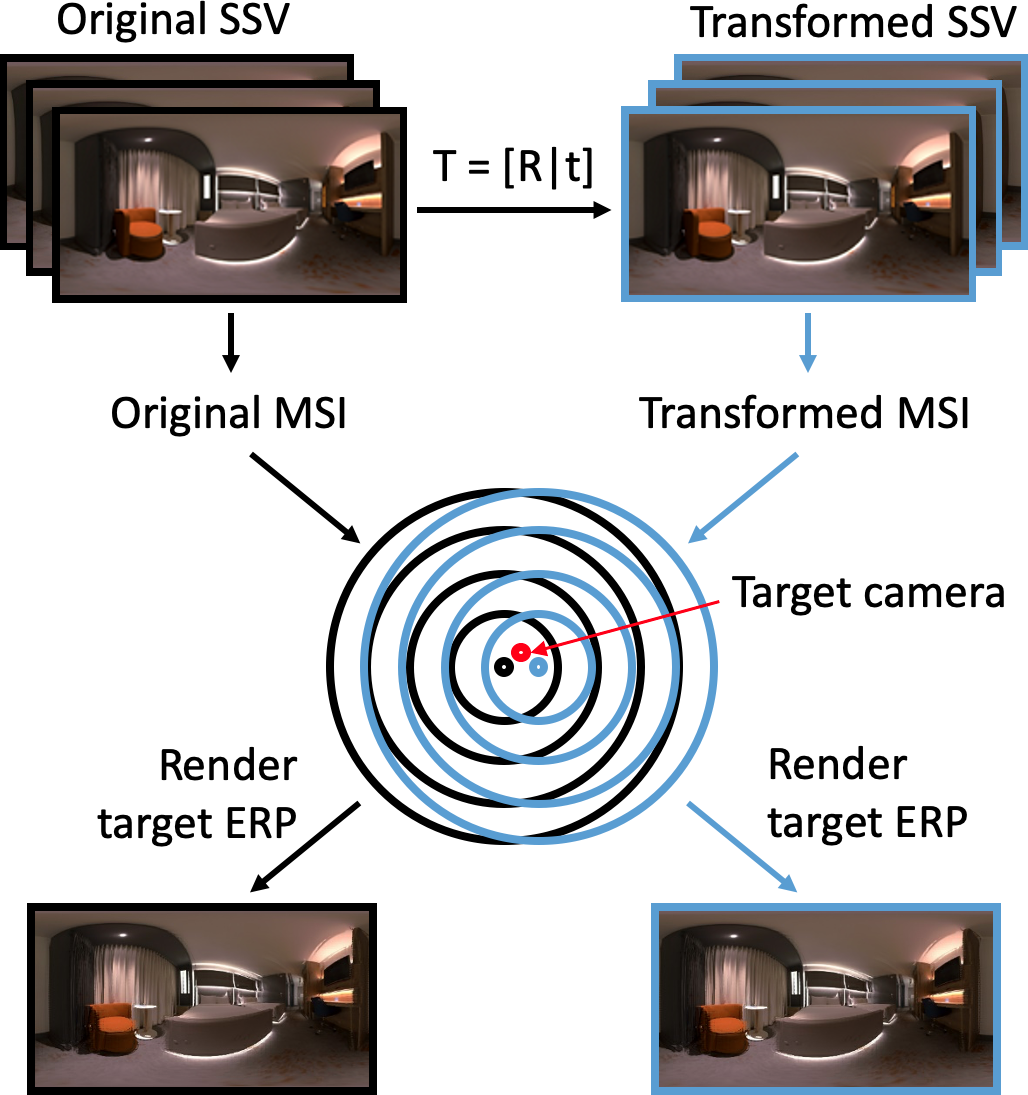}
\end{wrapfigure}
\noindent
Applying $T$ to the input corresponds to resampling the sphere sweeps for a set of concentric spheres transformed by $T$.
This is easily accomplished by applying $T$ to back-projected points in the sphere sweep volume generation process (\cref{sec:modelarchitecture}).
However, applying $T$ to an output MSI is less straightforward as, given an MSI, we must determine a new MSI at a pose transformed by $T$.
This %
requires interpolating alpha values and blending weights for the layers of the new MSI,
while still preserving the correct output color along rays, which may blur features of the original MSI.
Instead, our approach is to penalize the re-rendering loss for a target view with pose $P$ between the MSI predicted for transformed inputs, and the original MSI:
\begin{align}
	\label{eq:ti-rerender1}
    \mathcal{L}_\text{TI} = \norm{
        \render(f(T(\mathcal{S}^\text{L}), T(\mathcal{S}^\text{R})), P)
        - \render(T(f(\mathcal{S}^\text{L}, \mathcal{S}^\text{R})), P)
    }_2 \text{,}
\end{align}
which is equivalent to
\begin{align}
	\label{eq:ti-rerender2}
    \mathcal{L}_\text{TI} = \norm{
        \render(f(T(\mathcal{S}^\text{L}), T(\mathcal{S}^\text{R})), P)
        - \render(f(\mathcal{S}^\text{L}, \mathcal{S}^\text{R}), TP)
    }_2 \text{.}
\end{align}
	This can be computed without explicitly interpolating MSIs, and only requires transforming the target pose.
	If the re-renderings are close, then this implies the MSIs $f(T(\mathcal{S}_L), T(\mathcal{S}_R))$ and $T(f(\mathcal{S}_L, \mathcal{S}_R))$ are consistent scene representations.

\paragraph{Perceptual loss:}
Finally, we experiment with replacing the $\mathcal{L}_{\text{ERP-L2}}$ loss with a perceptual E-LPIPS~\cite{KettuHL2019} loss, which penalizes transformations of the image through comparisons of feature activations of the VGG network:
\begin{align}
    \mathcal{L}_\text{Per} &= \text{E-LPIPS}( \render(\text{MSI},P), I_\text{GT} ) \text{.}
\end{align}
Thus, our final loss is: $\cal{L}_{\text{Final}} = \lambda_\text{Per} \cal{L}_\text{Per} + \lambda_\text{TI} \cal{L}_\text{TI}$, with $\lambda_\text{Per}=1, \lambda_\text{TI}=10$. %

\subsection{High-resolution Rendering}

An advantage of any approach that directly predicts scene structure (MPIs, MSIs, meshes) is that, no matter the resolution of the output representation, it can be textured with a high-resolution image.
We combine \textit{multiple} high-resolution images (left and right ODS) using learned blending weights, which allows for high-resolution synthesis of both visible \textit{and} single-view occluded regions, while maintaining real-time frame rates.
We render a set of concentric spheres as meshes in Unity, textured with the alpha values described above and RGBs derived from combining the two high-resolution ODS images using the inferred low-resolution blending weights (\cref{eq:odsblend}).

Further, given an inferred MSI, we can apply a GPU-based joint bilateral upsampling filter \cite{KopfCLU2007} between the high-resolution 4K$\times$2K blended RGB images at each layer and the lower-resolution 640$\times$320 inferred alpha layers.
This offers the following advantages:
\begin{enumerate}
    \item It allows us to perform inference at low spatial resolution for speed, and then upsample the MSI alphas to match the higher-resolution input RGBs.
    \item Training produces checkerboard artifacts through the architecture of the strided transpose convolutions \cite{OdenaDO2016}; filtering reduces these, which improves the quality of occlusion edges during view synthesis.
\end{enumerate}
Different architecture choices are also possible to reduce checkerboarding, such as bilinear upsampling followed by convolution~\cite{SugawSK2018}, but typically these are more expensive on the GPU at inference time ($\approx2\times$ slower).

\section{Experiments}
\label{sec:experiments}

We experiment by training our model on a dataset of indoor scenes with moving cameras.
We quantitatively compare our model against ground-truth data with image quality and perceptual metrics.
Further, we test our approach on ODS footage from real cameras collected online---please see our supplemental video.

\paragraph{Training data:}
\label{sec:trainingdata}

Existing view synthesis and expansion approaches rely on large databases of permissively-licensed perspective video~\cite{ZhouTFFS2018}, which are not available for stereo 360\degree imagery as the format is still nascent.
Instead, we generate synthetic training data using the Replica dataset~\cite{StrauWMCWGEMRVCYBYPYZLCBGMPSBSNGLN2019} for supervision from \emph{target views}.
Replica contains high-quality scenes with few holes or missing/incorrect geometry, as might be found in real-world scan databases~\cite{ChangDFHNSSZZ2017,XiaZHSMS2018}.
In principle, we can use any projection for the target views during training (\cref{sec:projection}).
Choosing an omnidirectional projection allows us to back-propagate our loss through all parts of the MSI simultaneously.
If we trained on real video data, %
we would project inferred MSIs into target views by tracking the ODS video over time and selecting frames with desired poses.
However, our ultimate goal is to produce views that are correct to each human eye within head-tracked VR.
For this, ERP images more closely match our desire than ODS reprojections.
Given that our data is synthetic, we exploit this fact and render ERP target views for supervision.

\paragraph{Data generation:}

First, we develop a custom ERP and ODS renderer for Replica~\cite{StrauWMCWGEMRVCYBYPYZLCBGMPSBSNGLN2019}.
Then, we sample floor positions in Replica via uniformly randomly selecting from navigable positions available via the Facebook AI Habitat simulator~\cite{SavvaKMZWJSLKMPB2019}.
For each floor position, we sample a vertical position from a Gaussian distribution over human heights.
At each position, we render the camera's left-eye and right-eye ODS images, along with a triplet of ERP target views within the desired headbox for VR rendering. %
We render one \emph{interpolation} target view randomly within the ODS viewing circle, and two \emph{extrapolation} target views at random offset from the camera.
To reduce render aliasing as the synthetic data does not have mipmaps, we render each view at 4K$\times$2K, apply a Gaussian blur with $\sigma=3.2$\,pixels, and downsample to our training resolution of 640$\times$320.
We render ERP images from 30,000 positions in total.
Finally, we split the data 70\%/30\% across training and test sets by scene.
See our supplement for details.

\begin{table*}[t]
    \centering
    \caption{%
        \label{tab:resultstable-apt0}%
    	Quantitative comparison of baseline methods (top) and ablations of our approach (bottom).
    	We show single-image reconstruction error (E-LPIPS, SSIM, PSNR) on a Replica ODS test set.
    	We report 1000$\times$E-LPIPS \cite{KettuHL2019} as `mE-LPIPS' (milli-E-LPIPS).
    	For training data, (P) is perspective views and (ODS) is omnidirectional stereo. %
    	`\!\triup\!' means higher is better, `\tridown' means lower is better.
    	Numbers are mean$\pm$standard error.
    	Datasets: RealEstate10K \cite{ZhouTFFS2018}, Replica \cite{StrauWMCWGEMRVCYBYPYZLCBGMPSBSNGLN2019}, Stanford 2D-3D-S \cite{ArmenSZS2017}. %
        Best numbers in green.
    	*90\,Hz at 256$\times$128\,pixels $\approx$ 15\,Hz at 640$\times$320\,pixels. $^\dagger$Please see text discussion.
    }
    \resizebox{\textwidth}{!}{%
    \setlength{\tabcolsep}{3pt}%
    \newcommand{\best}[1]{\textcolor{green!70!black}{#1}}%
    \begin{tabular}{lrlccc} %
    	\toprule
    	\textbf{Baseline/Ablation Model}                &  \textbf{Inference\triup} & \textbf{Training Data} & \textbf{mE-LPIPS\tridown} & \textbf{SSIM\triup}  & \textbf{PSNR\triup}   \\
    	\midrule
    	Ground-truth depth+mesh rendering               &            N/A\phantom{*} & (not trained)          &       3.08$\pm$6.17       &    0.93$\pm$0.05     & 26.75$\pm$3.73        \\
    	Zhou et al.~\cite{ZhouTFFS2018} adapted to ODS  & 2\,Hz\phantom{*} & RealEstate10K (P)      &       7.38$\pm$4.00       &    0.78$\pm$0.06     & 20.65$\pm$2.15        \\
    	Double plane sweep (DPS-RE10K)                  &  \best{30\,Hz}\phantom{*} & RealEstate10K (P)      &       4.28$\pm$3.00       &    0.90$\pm$0.04     & 27.52$\pm$3.52        \\
    	Double plane sweep (DPS-Rep)                    &  \best{30\,Hz}\phantom{*} & Replica (P)            &       7.46$\pm$3.86       &    0.80$\pm$0.06     & 21.55$\pm$2.34        \\
    	ODS-Net~\cite{LaiXLL2019}+mesh rendering        &                   15\,Hz* & Stanford 2D-3D-S       &      5.58$\pm$3.54       &    0.82$\pm$0.06     & 21.82$\pm$3.06        \\
    	\midrule
    	Ours (real-time)                                &  \best{30\,Hz}\phantom{*} & Replica (ODS)          &       2.29$\pm$2.21       & \best{0.92$\pm$0.04} & 29.10$\pm$3.91        \\ %
    	\quad– CoordConv                                &  \best{30\,Hz}\phantom{*} & Replica (ODS)          &       2.43$\pm$2.10       & \best{0.92$\pm$0.04} & 29.14$\pm$3.92        \\ %
    	\qquad– E-LPIPS (using L2)                      &  \best{30\,Hz}\phantom{*} & Replica (ODS)          &       2.88$\pm$2.53       & \best{0.92$\pm$0.04} & 29.82$\pm$3.91        \\ %
    	Ours (with $2 \times 32 = 64$ MSI layers)       &         15\,Hz\phantom{*} & Replica (ODS)          &       2.45$\pm$2.25       & \best{0.92$\pm$0.04} & 29.25$\pm$3.89        \\ %
    	Ours (graph conv. network)                      &          2\,Hz\phantom{*} & Replica (ODS)          &       3.06$\pm$2.40       & \best{0.92$\pm$0.04} & \best{30.03$\pm$4.15} \\
    	Ours (with background layer)                    &          30\,Hz$^\dagger$ & Replica (ODS)          &   \best{2.04$\pm$2.14}    &    0.91$\pm$0.04     & \best{30.03$\pm$4.04} \\ %
    	\bottomrule
    \end{tabular}%
    }
\end{table*}

\paragraph{Image metrics:}

To quantitatively compare our results against ground truth and other baselines, we use three evaluation metrics: mean E-LPIPS~\cite{KettuHL2019}, SSIM~\cite{WangBSS2004}, and PSNR, along with their standard error over the test set of frames.
E-LPIPS is a relatively new metric, built upon LPIPS \cite{ZhangIESW2018}, which computes and compares VGG16 feature responses \cite{SimonZ2015} under different perturbations by self-ensembling through random transformations.
As a more advanced, learned `perceptual' metric, its intent is to be more robust to transformations which are equivalent in PSNR and SSIM but noticeable to the human eye~\cite{DosseY2011}.

\paragraph{Baseline Comparisons:}

To our knowledge, no existing method infers MSIs from 360\degree imagery;
we thus adapt existing baselines to our purpose and compare them in \cref{tab:resultstable-apt0}.
Our first baseline creates novel views using textured mesh rendering with ground-truth depth maps.
In practice, perfect depth is unattainable,
but it provides an upper bound on the performance of depth-based re-rendering.

\textbf{Zhou et al.’s}
released model \cite{ZhouUB2018} was trained on the extensive RealEstate10K dataset of perspective videos.
Their network takes as input a \emph{reference} view and a plane sweep volume for the \emph{source} view, and generates an MPI for the reference view.
We first naïvely extend their approach by providing the left ODS image as the reference view and a sphere sweep volume of the right ODS image as the source view.
This leads to bad performance across all metrics in \cref{tab:resultstable-apt0} and highlights the mismatch between Zhou et al.’s architecture and our desire to estimate an MSI at the center of the camera system from ODS imagery.

\textbf{Double-plane-sweep baseline.}
To better represent the nature of ODS imagery,
we create an adaption of Zhou et al.'s method that takes as input a plane sweep volume for each of the two views, and produces a multi-plane image.
We train this model on perspective views from the synthetic Replica and the natural RealEstate10K dataset. %
Please see the supplement for details.
Both datasets have similar content, but RealEstate10K is more varied and not synthetic.
At test time, we apply this model to sphere sweep volumes of the left/right ODS views, exploiting the `pseudo-perspective' projection in the equatorial region of ODS imagery.
Despite the mismatch between perspective training and ODS testing, the model trained on the RealEstate10K dataset performs well, and clearly better than the model that was trained on the less varied Replica dataset.

\textbf{ODS-Net~\cite{LaiXLL2019}}
is a real-time learning-based method specifically aimed at 6DoF video generation.
This method predicts a 256$\times$128 depth map per video frame, and synthesizes views by rendering a mesh based on the depth map (without inpainting).
Note that we provide double-ERP input, as expected by ODS-Net, while all other comparisons use ODS input.
In our experiments, ODS-Net failed to produce metrically accurate depth on our Replica test data, and performed worse than mesh-based rendering with ground-truth depth in \cref{tab:resultstable-apt0}.

\input{src/4-fig-serrano}

\textbf{Serrano et al.~\cite{SerraKCDGHM2019}}
is an offline optimization-based method for 6DoF video generation from RGBD ERP video input.
The method produces a set of RGBD+$\alpha$ layers for real-time rendering of novel views.
While their results show sharper occlusion boundaries and more accurate parallax, our method can better synthesize occluded content by extrapolating from both the left and right ODS images (see \cref{fig:serrano-comparison}).
As we do not assume that depth is available a priori, our method is also applicable to in-the-wild ODS videos without any additional preprocessing steps, such as depth estimation.
Our method also performs inference in real time at 30\,Hz, while Serrano et al.'s method requires one minute per frame.
Further, they assume a static camera for layer computation, while our method can, with sufficient training data, be applied to scenes with arbitrary camera motion.

\input{src/4-fig-perspective-puppy.tex}

\paragraph{Ablations:}

\cref{tab:resultstable-apt0} also shows quantitative measures for ablations of our approach, which reduce the important perceptual E-LPIPS score. %

\textbf{GCN.}
Our first ablation replaces convolutions directly on the ODS domain with a graph convolutional network (GCN) on a sphere mesh with approximately as many vertices as image pixels (164K).
We project the sphere sweep volumes into the spherical basis via this mesh, which then uses GCN layers (in architecture of Pixel2Mesh \cite{WangZLFLJ2018}) to transform them into per-vertex alpha/blending weights, which are projected back to ERP for the MSI.
This approach is (almost) rotation equivariant, and performs well in terms of the metrics in \cref{tab:resultstable-apt0}.
However, inference times are slow and currently cannot be accelerated to real-time with TensorRT due to its lack of sparse matrix support for graph convolutions.

\textbf{With background layer.}
We predict an additional RGB layer which can help to inpaint extrapolated disocclusions \cite{ZhouTFFS2018}.
This improves PSNR and E-LPIPS scores in \cref{tab:resultstable-apt0}.
However, at runtime, there are two issues to combine this with our high-resolution input videos:
1) the background layer is limited to the inference resolution, and so looks blurry in relation, and
2) we must explicitly find disoccluded regions to blend in this background layer, which complicates and slows virtual view render times. This is an application-level trade off.

\begin{table}[t]
    \caption{%
        \label{tab:temporaltable-apt0}%
    	Quantitative trade-off of transform-inverse regularization on temporal consistency and image quality, at $\lambda_{TI}=10$.
    	`f2f-(depth|rgb)' measures the average frame-to-frame difference between consecutive low-pass-filtered depth maps or RGB images as measure of temporal consistency.
    	`\!\triup\!' means higher is better, `\tridown' means lower is better.
    	Numbers are mean$\pm$standard error.
    `Transform range' indicates the scale factor on the transformed pose, where 1$\times$ is: translation $(x,y,z)\pm$0.01 metres; rotation $(\theta,\phi,\psi)\pm1.7$\degree.
    }
    \centering
    \setlength{\tabcolsep}{3pt}%
    \newcommand{\best}[1]{\textcolor{green!70!black}{#1}}%
    \begin{tabular}{ c c c c c c }
    	\toprule
    	\textbf{Transform range} & \textbf{f2f-depth\tridown} & \textbf{f2f-rgb\tridown} & \textbf{mE-LPIPS\tridown} & \textbf{SSIM\triup} & \textbf{PSNR\triup} \\
    	\midrule
    	None                     & 3.27$\pm$0.70              & 1.27$\pm$0.16            & 2.88$\pm$2.53             & 0.92$\pm$0.04       & 29.82$\pm$3.92      \\
    	$\times$0.5              & 1.64$\pm$0.37              & 1.29$\pm$0.16            & 2.37$\pm$2.18             & 0.92$\pm$0.04       & 29.74$\pm$4.07      \\
    	$\times$1.0              & 1.65$\pm$0.41              & 1.29$\pm$0.16            & 2.51$\pm$2.21             & 0.92$\pm$0.04       & 29.62$\pm$4.01      \\
    	$\times$2.0              & 1.48$\pm$0.28              & 1.29$\pm$0.16            & 2.59$\pm$2.27             & 0.92$\pm$0.04       & 29.52$\pm$4.19      \\
    	$\times$5.0              & 1.04$\pm$0.27              & 1.28$\pm$0.16            & 2.87$\pm$2.54             & 0.91$\pm$0.04       & 29.20$\pm$4.20      \\
    	\bottomrule
    \end{tabular}%
\end{table}

\textbf{Temporal consistency.}
We generate five ground-truth video sequences with moving cameras to measure temporal consistency.
Then, we apply a low-pass filter (with $\sigma=11$\,pixels) to the output videos, and compute absolute frame-to-frame differences (`f2f' metrics) between consecutive frames and depth maps, which detects temporal inconsistencies such as flickering.
As we increase the transformation range for transform-inverse regularization, f2f RGB metrics improve while image metrics degrade slightly due to increased blur in the output MSI RGBs and alphas.
Beyond RGB, depth consistency improves more significantly---this is important since we desire accurate and consistent re-rendering and disparity cues for consecutive frames, and for different eye IPDs in VR.

\input{src/4-fig-puppy-msi.tex}
\input{src/4-fig-puppy-w-odsnet}

\paragraph{Qualitative results:}

First, we show qualitative comparisons for reprojected views against Zhou et al.'s method~\cite{ZhouTFFS2018} augmented with a double-plane-sweep input architecture (\cref{fig:puppycomparebaselineods}). %
These use our real-time pipeline with high-resolution input videos applied to the MSI.
Our approach produces results which are visually sharper as we extrapolate the virtual camera away from the baseline at $3\times$ and $5\times$ interpupillary distance (IPD).
In \cref{fig:results1a,fig:results1b}, we show the output of our approach at the inference resolution of 640$\times$320, i.e., without the full real-time pipeline which uses the high-resolution input video. %
We generate the MSI representation at the center of the ODS camera system, and then generate ODS imagery at 1$\times$ IPD to show reconstruction, and at 3$\times$ and 5$\times$ IPD to show extrapolation (a similar max. distance to Zhou et al.\ \cite{ZhouTFFS2018}).
In our supplement, we show results on synthetic and real ODS video test sequences.

\section{Discussion and Conclusion}

Via the MSI representation and our learned network, our approach provides real-time inference on ODS video, and stereo VR rendering at 80\,Hz.
On our test set, we demonstrate results on baselines up to 5.6$\times$ larger than ODS interpupillary distance, and produce the highest quantitative performance.
In a VR headset, the user's motion is unconstrained and can lead to \emph{much} larger baselines than any current technique can handle~\cite{ChoiGTKK2019,SriniTBRNS2019}, especially in real-time.
The effect is to `see behind the curtain': our representation still provides correct disparity and motion parallax for scene objects with accurate alpha (which improves comfort), but large disocclusions reveal the layered MSI structure underneath
(see supplement).

Some multi-camera systems can see more scene content than is captured in the final ODS projection, and designing a system to start from raw camera feeds would allow this content to contribute to a reconstruction.
Our proposed approach could be adapted to work with per-camera feeds, which we leave for future work.
Our approach is applicable to online ‘in the wild’ ODS videos as well as ODS live streams supported by several off-the-shelf cameras.

\paragraph{Limitations:}

The quality at larger extrapolations is currently limited by computational trade-offs both during training and in real-time application.
First, our per-layer ODS blending for re-rendering could be improved by flow-based interpolation methods\ \cite{BerteCR2019}, which would be required during both training and inference.
Second, inferring higher-resolution MSI representations is possible at a slower speed with a network with more layers \cite{FlynnBDDFOST2019,SriniTBRNS2019,BroxtFOEHDDBWD2020} and so a larger receptive field.
Third, an explicit spherical convolution approach may improve quality, but current approaches are too expensive for real-time applications.
Further, stereo 360\degree video is a nascent format, and there is insufficient training data available.
Given our training data, the proposed method works best on indoor scenes with moving cameras, and more work must be done for dynamic scene objects. %

MSI depth discretization limits the range of non-aliased views, beyond which the layered nature of the MSI becomes noticeable~\cite{SriniTBRNS2019}. Increasing only the spatial resolution of the RGB for each MSI layer via the high-resolution input video increases this effect. However, this approach to VR rendering can be a better trade-off in terms of quality and comfort than keeping low-resolution imagery.

\paragraph{Conclusion:}

Stereo 360\degree video is only `half-way' to comfortable (and thus useful) representations of our environments, with the missing piece being 6DoF video.
Our work suggests one solution to this problem by learning to create representations which implicitly compute depth and fill disoccluded regions.
Our end-to-end system takes images from a stereo 360\degree camera and converts them into a 6DoF multi-sphere image representation in real time for viewing.
This learns how to distribute the scene over different depths per frame, and ensures temporal consistency.
We show competitive quantitative metrics for image quality while remaining fast in inference speed---this is important for situations where preprocessing is not an option, like communications and robotic teleoperation. Overall, we move towards a more natural 6DoF viewing experience for stereo 360\degree video.
\looseness=-1

\paragraph{Acknowledgments:}
We thank Ana Serrano for help with RGB-D comparisons and Eliot Laidlaw for improving the Unity renderer.
We thank Frédéric Devernay, Brian Berard, and an Amazon Research Award, and NVIDIA for a GPU donation.
This work was supported by a Brown OVPR Seed Award, RCUK grant CAMERA (EP/M023281/1), and an EPSRC-UKRI Innovation Fellowship (EP/S001050/1).

\bibliographystyle{splncs04}
{
\renewcommand{\bibname}{References}
\renewcommand{\bibsection}{{\section*{\bibname}}}
\small
\bibliography{bib/360-view-expansion,bib/biblio}{}

\begin{thebibliography}{10}
\providecommand{\url}[1]{\texttt{#1}}
\providecommand{\urlprefix}{URL }
\providecommand{\doi}[1]{https://doi.org/#1}

\bibitem{AnderGBKSHAS2016}
Anderson, R., Gallup, D., Barron, J.T., Kont\-kanen, J., Snavely, N.,
  Hernandez, C., Agarwal, S., Seitz, S.M.: Jump: Virtual reality video. ACM
  Trans. Graph.  \textbf{35}(6),  198:1--13 (2016).
  \doi{10.1145/2980179.2980257}

\bibitem{ArmenSZS2017}
Armeni, I., Sax, S., Zamir, A.R., Savarese, S.: Joint {{2D}-3D}-semantic data
  for indoor scene understanding (2017),
  \href{https://arxiv.org/abs/1702.01105}{arXiv:1702.01105}

\bibitem{BerteCR2019}
Bertel, T., Campbell, N.D.F., Richardt, C.: {MegaParallax}: Casual 360°
  panoramas with motion parallax. TVCG  \textbf{25}(5),  1828--1835 (2019).
  \doi{10.1109/TVCG.2019.2898799}

\bibitem{BonneTSSPP2015}
Bonneel, N., Tompkin, J., Sunkavalli, K., Sun, D., Paris, S., Pfister, H.:
  Blind video temporal consistency. ACM Trans. Graph.  \textbf{34}(6),
  196:1--9 (2015). \doi{10.1145/2816795.2818107}

\bibitem{BrownL2007}
Brown, M., Lowe, D.G.: Automatic panoramic image stitching using invariant
  features. IJCV  \textbf{74}(1),  59--73 (2007).
  \doi{10.1007/s11263-006-0002-3}

\bibitem{BroxtFOEHDDBWD2020}
Broxton, M., Flynn, J., Overbeck, R., Erickson, D., Hedman, P., DuVall, M.,
  Dourgarian, J., Busch, J., Whalen, M., Debevec, P.: Immersive light field
  video with a layered mesh representation. ACM Trans. Graph.  \textbf{39}(4),
  86:1--15 (2020). \doi{10.1145/3386569.3392485}

\bibitem{ChangDFHNSSZZ2017}
Chang, A., Dai, A., Funkhouser, T., Halber, M., Nießner, M., Savva, M., Song,
  S., Zeng, A., Zhang, Y.: {Matterport3D}: Learning from {RGB-D} data in indoor
  environments. In: 3DV. pp. 667--676 (2017). \doi{10.1109/3DV.2017.00081}

\bibitem{ChapdR2013}
Chapdelaine-Couture, V., Roy, S.: The omnipolar camera: A new approach to
  stereo immersive capture. In: ICCP (2013). \doi{10.1109/ICCPhot.2013.6528311}

\bibitem{ChengCDWLS2018}
Cheng, H.T., Chao, C.H., Dong, J.D., Wen, H.K., Liu, T.L., Sun, M.: Cube
  padding for weakly-supervised saliency prediction in 360° videos. In: CVPR.
  pp. 1420--1429 (2018). \doi{10.1109/CVPR.2018.00154}

\bibitem{ChoiGTKK2019}
Choi, I., Gallo, O., Troccoli, A., Kim, M.H., Kautz, J.: Extreme view
  synthesis. In: ICCV. pp. 7780--7789 (2019). \doi{10.1109/ICCV.2019.00787}

\bibitem{CohenGKW2018}
Cohen, T.S., Geiger, M., Koehler, J., Welling, M.: Spherical {CNNs}. In: ICLR
  (2018)

\bibitem{CoorsCG2018}
Coors, B., Condurache, A.P., Geiger, A.: {SphereNet}: Learning spherical
  representations for detection and classification in omnidirectional images.
  In: ECCV. pp. 518--533 (2018). \doi{10.1007/978-3-030-01240-3_32}

\bibitem{Dodgs2004}
Dodgson, N.A.: Variation and extrema of human interpupillary distance. In:
  Stereoscopic Displays and Virtual Reality Systems (2004).
  \doi{10.1117/12.529999}

\bibitem{DosseY2011}
Dosselmann, R., Yang, X.D.: A comprehensive assessment of the structural
  similarity index. Signal, Image and Video Processing  \textbf{5}(1),  81--91
  (2011). \doi{10.1007/s11760-009-0144-1}

\bibitem{EilerMU2019}
Eilertsen, G., Mantiuk, R.K., Unger, J.: Single-frame regularization for
  temporally stable {CNNs}. In: CVPR. pp. 11168--11177 (2019).
  \doi{10.1109/CVPR.2019.01143}

\bibitem{EstevAMD2018}
Esteves, C., Allen-Blanchette, C., Makadia, A., Daniilidis, K.: Learning
  {SO(3)} equivariant representations with spherical {CNNs}. In: ECCV. pp.
  52--68 (2018). \doi{10.1007/978-3-030-01261-8_4}

\bibitem{FlynnBDDFOST2019}
Flynn, J., Broxton, M., Debevec, P., DuVall, M., Fyffe, G., Overbeck, R.,
  Snavely, N., Tucker, R.: {DeepView}: View synthesis with learned gradient
  descent. In: CVPR. pp. 2367--2376 (2019). \doi{10.1109/CVPR.2019.00247}

\bibitem{Googl2015}
{Google Inc.}: Rendering omni-directional stereo content (2015),
  \url{https://developers.google.com/vr/jump/rendering-ods-content.pdf}

\bibitem{HuangCCJ2017}
Huang, J., Chen, Z., Ceylan, D., Jin, H.: 6-{DOF VR} videos with a single
  360-camera. In: IEEE VR. pp. 37--44 (2017). \doi{10.1109/VR.2017.7892229}

\bibitem{ImHRJCK2016}
Im, S., Ha, H., Rameau, F., Jeon, H.G., Choe, G., Kweon, I.S.: All-around depth
  from small motion with a spherical panoramic camera. In: ECCV (2016).
  \doi{10.1007/978-3-319-46487-9_10}

\bibitem{IshigYT1992}
Ishiguro, H., Yamamoto, M., Tsuji, S.: Omni-directional stereo. TPAMI
  \textbf{14}(2),  257--262 (1992). \doi{10.1109/34.121792}

\bibitem{KettuHL2019}
Kettunen, M., Härkönen, E., Lehtinen, J.: {E-LPIPS}: Robust perceptual image
  similarity via random transformation ensembles (2019),
  \href{https://arxiv.org/abs/1906.03973}{arXiv:1906.03973}

\bibitem{KonraDMW2017}
Konrad, R., Dansereau, D.G., Masood, A., Wetzstein, G.: {SpinVR}: Towards
  live-streaming {3D} virtual reality video. ACM Trans. Graph.  \textbf{36}(6),
   209:1--12 (2017). \doi{10.1145/3130800.3130836}

\bibitem{Kopf2016}
Kopf, J.: 360° video stabilization. ACM Trans. Graph.  \textbf{35}(6),
  195:1--9 (2016). \doi{10.1145/2980179.2982405}

\bibitem{KopfCLU2007}
Kopf, J., Cohen, M.F., Lischinski, D., Uyttendaele, M.: Joint bilateral
  upsampling. ACM Trans. Graph.  \textbf{26}(3), ~96 (2007).
  \doi{10.1145/1276377.1276497}

\bibitem{LaiXLL2019}
Lai, P.K., Xie, S., Lang, J., Laganière, R.: Real-time panoramic depth maps
  from omni-directional stereo images for 6 {DoF} videos in virtual reality.
  In: IEEE VR. pp. 405--412 (2019). \doi{10.1109/VR.2019.8798016}

\bibitem{LaiHWSYY2018}
Lai, W.S., Huang, J.B., Wang, O., Shechtman, E., Yumer, E., Yang, M.H.:
  Learning blind video temporal consistency. In: ECCV. pp. 170--185 (2018).
  \doi{10.1007/978-3-030-01267-0_11}

\bibitem{LeeKKKN2016}
Lee, J., Kim, B., Kim, K., Kim, Y., Noh, J.: Rich360: Optimized spherical
  representation from structured panoramic camera arrays. ACM Trans. Graph.
  \textbf{35}(4),  63:1--11 (2016). \doi{10.1145/2897824.2925983}

\bibitem{LeeJYJY2019}
Lee, Y.K., Jeong, J., Yun, J.S., June, C.W., Yoon, K.J.: {SpherePHD}: Applying
  {CNNs} on a spherical {PolyHeDron} representation of 360° images. In: CVPR.
  pp. 9173--9181 (2019). \doi{10.1109/CVPR.2019.00940}

\bibitem{LiuLMSFSY2018}
Liu, R., Lehman, J., Molino, P., Such, F.P., Frank, E., Sergeev, A., Yosinski,
  J.: An intriguing failing of convolutional neural networks and the
  {CoordConv} solution. In: NeurIPS (2018)

\bibitem{LuoXRY2018}
Luo, B., Xu, F., Richardt, C., Yong, J.H.: Parallax360: Stereoscopic 360°
  scene representation for head-motion parallax. TVCG  \textbf{24}(4),
  1545--1553 (2018). \doi{10.1109/TVCG.2018.2794071}

\bibitem{MatzeCEKS2017}
Matzen, K., Cohen, M.F., Evans, B., Kopf, J., Szeliski, R.: Low-cost 360 stereo
  photography and video capture. ACM Trans. Graph.  \textbf{36}(4),  148:1--12
  (2017). \doi{10.1145/3072959.3073645}

\bibitem{MildeSOKRNK2019}
Mildenhall, B., Srinivasan, P.P., Ortiz-Cayon, R., Kalantari, N.K.,
  Ramamoorthi, R., Ng, R., Kar, A.: Local light field fusion: Practical view
  synthesis with prescriptive sampling guidelines. ACM Trans. Graph.
  \textbf{38}(4),  29:1--14 (2019). \doi{10.1145/3306346.3322980}

\bibitem{OdenaDO2016}
Odena, A., Dumoulin, V., Olah, C.: Deconvolution and checkerboard artifacts.
  Distill  (2016). \doi{10.23915/distill.00003}

\bibitem{PadmaRSNW2018}
Padmanaban, N., Ruban, T., Sitzmann, V., Norcia, A.M., Wetzstein, G.: Towards a
  machine-learning approach for sickness prediction in 360° stereoscopic
  videos. TVCG  \textbf{24}(4),  1594--1603 (2018).
  \doi{10.1109/TVCG.2018.2793560}

\bibitem{ParraTFHMSSC2019}
Parra~Pozo, A., Toksvig, M., Filiba~Schrager, T., Hsu, J., Mathur, U.,
  Sorkine-Hornung, A., Szeliski, R., Cabral, B.: An integrated {6DoF} video
  camera and system design. ACM Trans. Graph.  \textbf{38}(6),  216:1--16
  (2019). \doi{10.1145/3355089.3356555}

\bibitem{PelegBP2001}
Peleg, S., Ben-Ezra, M., Pritch, Y.: Omnistereo: Panoramic stereo imaging.
  TPAMI  \textbf{23}(3),  279--290 (2001). \doi{10.1109/34.910880}

\bibitem{PenneZ2017}
Penner, E., Zhang, L.: Soft {3D} reconstruction for view synthesis. ACM Trans.
  Graph.  \textbf{36}(6),  235:1--11 (2017). \doi{10.1145/3130800.3130855}

\bibitem{PerazSZKWWG2015}
Perazzi, F., Sorkine-Hornung, A., Zimmer, H., Kaufmann, P., Wang, O., Watson,
  S., Gross, M.: Panoramic video from unstructured camera arrays. Comput.
  Graph. Forum  \textbf{34}(2),  57--68 (2015). \doi{10.1111/cgf.12541}

\bibitem{PorteD1984}
Porter, T., Duff, T.: Compositing digital images. Computer Graphics
  (Proceedings of SIGGRAPH)  \textbf{18}(3),  253--259 (1984).
  \doi{10.1145/800031.808606}

\bibitem{Richa2020}
Richardt, C.: Omnidirectional stereo. In: Ikeuchi, K. (ed.) Computer Vision: A
  Reference Guide. Springer (2020). \doi{10.1007/978-3-030-03243-2_808-1}

\bibitem{RichaPZS2013}
Richardt, C., Pritch, Y., Zimmer, H., Sorkine-Hornung, A.: Megastereo:
  Constructing high-resolution stereo panoramas. In: CVPR. pp. 1256--1263
  (2013). \doi{10.1109/CVPR.2013.166}

\bibitem{RichaTHHSW2017}
Richardt, C., Tompkin, J., Halsey, J., Hertzmann, A., Starck, J., Wang, O.:
  Video for virtual reality. In: SIGGRAPH Courses (2017).
  \doi{10.1145/3084873.3084894}

\bibitem{RonneFB2015}
Ronneberger, O., Fischer, P., Brox, T.: {U-Net}: Convolutional networks for
  biomedical image segmentation. In: MICCAI. pp. 234--241 (2015).
  \doi{10.1007/978-3-319-24574-4_28}

\bibitem{SavvaKMZWJSLKMPB2019}
Savva, M., Kadian, A., Maksymets, O., Zhao, Y., Wijmans, E., Jain, B., Straub,
  J., Liu, J., Koltun, V., Malik, J., Parikh, D., Batra, D.: {Habitat}: A
  platform for embodied {AI} research. In: ICCV. pp. 9339--9347 (2019).
  \doi{10.1109/ICCV.2019.00943}

\bibitem{SchroBS2018}
Schroers, C., Bazin, J.C., Sorkine-Hornung, A.: An omnistereoscopic video
  pipeline for capture and display of real-world {VR}. ACM Trans. Graph.
  \textbf{37}(3),  37:1--13 (2018). \doi{10.1145/3225150}

\bibitem{SerraKCDGHM2019}
Serrano, A., Kim, I., Chen, Z., DiVerdi, S., Gutierrez, D., Hertzmann, A.,
  Masia, B.: Motion parallax for 360° {RGBD} video. TVCG  \textbf{25}(5),
  1817--1827 (2019). \doi{10.1109/TVCG.2019.2898757}

\bibitem{SimonZ2015}
Simonyan, K., Zisserman, A.: Very deep convolutional networks for large-scale
  image recognition. In: ICLR (2015)

\bibitem{SriniTBRNS2019}
Srinivasan, P.P., Tucker, R., Barron, J.T., Ramamoorthi, R., Ng, R., Snavely,
  N.: Pushing the boundaries of view extrapolation with multiplane images. In:
  CVPR. pp. 175--184 (2019). \doi{10.1109/CVPR.2019.00026}

\bibitem{StrauWMCWGEMRVCYBYPYZLCBGMPSBSNGLN2019}
Straub, J., Whelan, T., Ma, L., Chen, Y., Wijmans, E., Green, S., Engel, J.J.,
  Mur-Artal, R., Ren, C., Verma, S., Clarkson, A., Yan, M., Budge, B., Yan, Y.,
  Pan, X., Yon, J., Zou, Y., Leon, K., Carter, N., Briales, J., Gillingham, T.,
  Mueggler, E., Pesqueira, L., Savva, M., Batra, D., Strasdat, H.M., Nardi,
  R.D., Goesele, M., Lovegrove, S., Newcombe, R.: The {Replica} dataset: A
  digital replica of indoor spaces (2019),
  \href{https://arxiv.org/abs/1906.05797}{arXiv:1906.05797}

\bibitem{SuG2017a}
Su, Y.C., Grauman, K.: Learning spherical convolution for fast features from
  360° imagery. In: NIPS (2017)

\bibitem{SuG2019}
Su, Y.C., Grauman, K.: Kernel transformer networks for compact spherical
  convolution. In: CVPR. pp. 9442--9451 (2019). \doi{10.1109/CVPR.2019.00967}

\bibitem{SugawSK2018}
Sugawara, Y., Shiota, S., Kiya, H.: Super-resolution using convolutional neural
  networks without any checkerboard artifacts. In: ICIP. pp. 66--70 (2018).
  \doi{10.1109/ICIP.2018.8451141}

\bibitem{Szeli2006}
Szeliski, R.: Image alignment and stitching: a tutorial. Foundations and Trends
  in Computer Graphics and Vision  \textbf{2}(1),  1--104 (2006).
  \doi{10.1561/0600000009}

\bibitem{TatenNT2018}
Tateno, K., Navab, N., Tombari, F.: Distortion-aware convolutional filters for
  dense prediction in panoramic images. In: ECCV. pp. 732--750 (2018).
  \doi{10.1007/978-3-030-01270-0_43}

\bibitem{ThattBLG2016}
Thatte, J., Boin, J.B., Lakshman, H., Girod, B.: Depth augmented stereo
  panorama for cinematic virtual reality with head-motion parallax. In: ICME
  (2016). \doi{10.1109/ICME.2016.7552858}

\bibitem{ThattG2018}
Thatte, J., Girod, B.: Towards perceptual evaluation of six degrees of freedom
  virtual reality rendering from stacked {OmniStereo} representation.
  Electronic Imaging  \textbf{2018}(5),  352--1--6 (2018).
  \doi{10.2352/ISSN.2470-1173.2018.05.PMII-352}

\bibitem{WangHCLYSCS2018}
Wang, F.E., Hu, H.N., Cheng, H.T., Lin, J.T., Yang, S.T., Shih, M.L., Chu,
  H.K., Sun, M.: Self-supervised learning of depth and camera motion from 360°
  videos. In: ACCV. pp. 53--68 (2018). \doi{10.1007/978-3-030-20873-8_4}

\bibitem{WangZLFLJ2018}
Wang, N., Zhang, Y., Li, Z., Fu, Y., Liu, W., Jiang, Y.G.: {Pixel2Mesh}:
  Generating {3D} mesh models from single {RGB} images. In: ECCV. pp. 52--67
  (2018). \doi{10.1007/978-3-030-01252-6_4}

\bibitem{WangBSS2004}
Wang, Z., Bovik, A.C., Sheikh, H.R., Simoncelli, E.P.: Image quality
  assessment: from error visibility to structural similarity. IEEE Transactions
  on Image Processing  \textbf{13}(4),  600--612 (2004).
  \doi{10.1109/TIP.2003.819861}

\bibitem{XiaZHSMS2018}
Xia, F., Zamir, A.R., He, Z., Sax, A., Malik, J., Savarese, S.: {Gibson Env}:
  Real-world perception for embodied agents. In: CVPR. pp. 9068--9079 (2018).
  \doi{10.1109/CVPR.2018.00945}

\bibitem{ZhangLSC2019}
Zhang, C., Liwicki, S., Smith, W., Cipolla, R.: Orientation-aware semantic
  segmentation on icosahedron spheres. In: ICCV. pp. 3533--3541 (2019).
  \doi{10.1109/ICCV.2019.00363}

\bibitem{ZhangIESW2018}
Zhang, R., Isola, P., Efros, A.A., Shechtman, E., Wang, O.: The unreasonable
  effectiveness of deep features as a perceptual metric. In: CVPR (2018).
  \doi{10.1109/CVPR.2018.00068}

\bibitem{ZhouUB2018}
Zhou, H., Ummenhofer, B., Brox, T.: {DeepTAM}: Deep tracking and mapping. In:
  ECCV. pp. 822--838 (2018). \doi{10.1007/978-3-030-01270-0_50}

\bibitem{ZhouTFFS2018}
Zhou, T., Tucker, R., Flynn, J., Fyffe, G., Snavely, N.: Stereo magnification:
  Learning view synthesis using multiplane images. ACM Trans. Graph.
  \textbf{37}(4),  65:1--12 (2018). \doi{10.1145/3197517.3201323}

\bibitem{ZioulKZAD2019}
Zioulis, N., Karakottas, A., Zarpalas, D., Alvarez, F., Daras, P.: Spherical
  view synthesis for self-supervised 360° depth estimation. In: 3DV. pp.
  690--699 (2019). \doi{10.1109/3DV.2019.00081}

\bibitem{ZioulKZD2018}
Zioulis, N., Karakottas, A., Zarpalas, D., Daras, P.: {OmniDepth}: Dense depth
  estimation for indoors spherical panoramas. In: ECCV. pp. 448--465 (2018).
  \doi{10.1007/978-3-030-01231-1_28}

\end{thebibliography}
}

\newpage

\appendix
\section*{\centering{MatryODShka: Appendices}}
\addcontentsline{toc}{section}{Appendices}
\renewcommand{\thesubsection}{\Alph{subsection}}

\noindent
These appendices contain
additional results and comparisons (\cref{sec:additionalresults}) as well as implementation details of our approach, including
our used hardware and software (\cref{sec:hardware-software}),
our joint bilateral upsampling (\cref{sec:jbu}),
details of our architecture and hyperparameters (\cref{sec:networkarchitecture}), and
rendering pseudocode (\cref{sec:rendering}).

\subsection{Additional Results and Comparisons}
\label{sec:additionalresults}

We show additional results and comparisons in \cref{fig:cafeteria-msi,fig:cafeteria-ipd,fig:mammamia-msi,fig:mammamia-ipd,fig:grandcanyon-msi,fig:grandcanyon-ipd}, and in video form in our supplemental materials.
This includes two additional MSI decompositions in \cref{fig:cafeteria-msi,fig:mammamia-msi,fig:grandcanyon-msi}, and view synthesis comparisons to the perspective double-plane-sweep baseline in \cref{fig:cafeteria-ipd,fig:mammamia-ipd,fig:grandcanyon-ipd}.
Our approach produces better novel views at larger synthesis baselines than the baseline method.
We illustrate the limitations of a layered representation such as ours in \cref{fig:limitations}.
Please see our supplemental video for additional results.

\input{src/4-fig-cafeteria-msi.tex}
\input{src/4-fig-cafeteria-ipd.tex}

\input{src/4-fig-mammamia-msi.tex}
\input{src/4-fig-mammamia-ipd.tex}

\input{src/4-fig-grandcanyon-msi.tex}
\input{src/4-fig-grandcanyon-ipd.tex}

\begin{figure}[t]
	\centering
	\includegraphics[height=7cm]{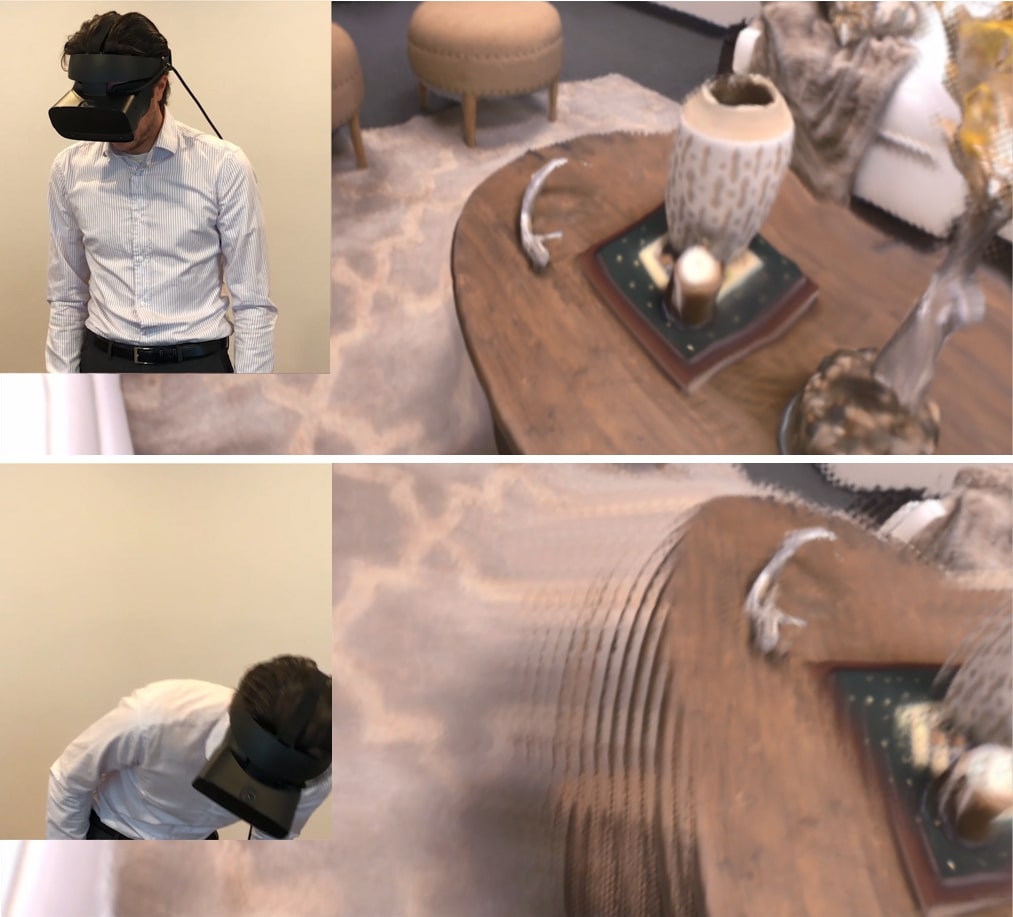}
	\caption{\emph{Limitation:} Using our system within a VR headset allows large motions away from the center of the MSI, exposing the layer structure of the representation (\emph{bottom}).}
	\label{fig:limitations}
\end{figure}

\clearpage
\subsection{Implementation Details}
\label{sec:additionaltechnicaldetails}

\subsubsection{Hardware and Software}
\label{sec:hardware-software}

Simultaneously decoding high-resolution video (e.g., 4K$\times$4K at 30\,Hz), inferring MSIs, and rendering stereo video from MSIs into a VR headset requires significant compute. 
For the headset, we use an Oculus Rift S, which requires rendering at 80\,Hz.
We compare two current PC platforms: a current discrete-GPU laptop and a workstation.
The laptop has an Intel(R) Core(TM) i7-8750H CPU with 16\,GB RAM, and NVIDIA GeForce RTX 2080 with Max-Q Design. This provides 30\,Hz MSI inference and 60+\,Hz novel-view rendering (30+\,Hz in VR). This rendering speed is sufficient for general desktop viewing, but not sufficient for fast head motion in VR.
The workstation has an AMD Ryzen 2950X CPU with 64\,GB RAM, and two NVIDIA GeForce 2080 Ti GPUs with RTX bridge to allow fast inter-GPU memory copy. This provides 30\,Hz MSI inference and 250+\,Hz novel-view rendering (125+\,Hz in VR).

We train our model using TensorFlow via our TensorFlow-based reprojection renderer. 
For our inference engine within our Unity-based renderer, we use TensorRT for efficient GPU computation.
We convert our model weights to 16-bit floating-point precision and load the model in TensorRT using ONNX. 
Further, we use CUDA for anti-aliased video downsampling to the network's expected input resolution, and for ODS sphere sweep volume creation. 
Finally, we implement ODS reprojection rendering from our MSIs, and our joint-bilateral upsampling, using Unity's CG shaders.

\subsubsection{Joint-Bilateral Upsampling Effect}
\label{sec:jbu}

Our network architecture uses learned upsampling within its U-Net via transpose convolution layers, which is known to introduce checkerboarding artifacts but is approximately 2$\times$ faster to infer than bilinear upsampling followed by learned convolution~\cite{SugawSK2018}.
To correct for these artifacts, we use joint-bilateral upsampling \cite{Kopf2016} on the screen-space perspective view as we accumulate the alpha layers and blend the RGB inputs within our real-time renderer.
This successfully removes some artifacts.

Bilateral filters are computationally expensive, yet this approach is possible because of our combined hardware and software design: 1) We use a dedicated GPU for inference, which we wish to run as fast as possible for real-time video,
and so we make a trade off in the quality of the model inference because 2) We use a dedicated GPU for rendering; rendering the multi-sphere representation is fast, and so we have spare compute capacity on the render card for this filtering.
In a setting with a less powerful machine, the filtering could be skipped.

\subsubsection{Network Architecture and Hyperparameters}
\label{sec:networkarchitecture}

We include a complete layer description of our network architecture in \cref{tab:architecture}.
We also include a table of all of our architecture and system hyperparameters (\cref{tab:parameters}).

\begin{table}[tp]
\newcommand{\plusone}{\textcolor{green!50!black}{+1}}
\caption{U-Net-style convolutional neural network architecture for our approach, as shown in Figure 2 of the main paper. `\textbf{k}' is the kernel size,
`\textbf{s}' is the stride,
`\textbf{d}' is the dilation factor of the kernel,
and `\textbf{chns}' is the number of input/output channels (kernels).
The network input are the left and right sphere sweep volumes $\mathcal{S}^\text{L}$ and $\mathcal{S}^\text{R}$. The internal convolutional layers are identical to that of the architecture in \cite{ZhouTFFS2018}, except that input to each convolutional layer is concatenated with the $V$ coordinate (`\plusone' channel), as described in the main paper. As in \cite{ZhouTFFS2018}, each convolutional layer is followed by layer normalization and ReLU non-linearity, except for the last layer.
The double-plane sweep baseline uses the same architecture, without additional coordinate channels.}
\label{tab:architecture}
\centering
\setlength{\tabcolsep}{6pt}%
\begin{tabular}{c*{7}{c}}
\toprule
\textbf{Layer} & \textbf{k} & \textbf{s} & \textbf{d} & \textbf{chns} & \textbf{in} & \textbf{out} & \textbf{input} \\
\midrule
conv1\_1 & 3 & 1 & 1 &  2$\times$32$\times$3\plusone/64  & 1 & 1 & $\mathcal{S}^\text{L}$ + $\mathcal{S}^\text{R}$ + $V$ \\
conv1\_2 & 3 & 2 & 1 &  64\plusone/128 & 1 & 2 & conv1\_1 + $V$ \\
conv2\_1 & 3 & 1 & 1 & 128\plusone/128 & 2 & 2 & conv1\_2 + $V$ \\
conv2\_2 & 3 & 2 & 1 & 128\plusone/256 & 2 & 4 & conv2\_1 + $V$ \\
conv3\_1 & 3 & 1 & 1 & 256\plusone/256 & 4 & 4 & conv2\_2 + $V$ \\
conv3\_2 & 3 & 1 & 1 & 256\plusone/256 & 4 & 4 & conv3\_1 + $V$ \\
conv3\_3 & 3 & 2 & 1 & 256\plusone/512 & 4 & 8 & conv3\_2 + $V$ \\
conv4\_1 & 3 & 1 & 2 & 512\plusone/512 & 8 & 8 & conv3\_3 + $V$ \\
conv4\_2 & 3 & 1 & 2 & 512\plusone/512 & 8 & 8 & conv4\_1 + $V$ \\
conv4\_3 & 3 & 1 & 2 & 512\plusone/512 & 8 & 8 & conv4\_2 + $V$ \\
\midrule
conv5\_1 & 4 & .5 & 1 & 1024\plusone/256 & 8 & 4 & conv4\_3 + conv3\_3 + $V$ \\
conv5\_2 & 3 &  1 & 1 &  256\plusone/256 & 4 & 4 & conv5\_1 + $V$ \\
conv5\_3 & 3 &  1 & 1 &  256\plusone/256 & 4 & 4 & conv5\_2 + $V$ \\
conv6\_1 & 4 & .5 & 1 &  512\plusone/128 & 4 & 2 & conv5\_3 + conv2\_2 + $V$ \\
conv6\_2 & 3 &  1 & 1 &  128\plusone/128 & 2 & 2 & conv6\_1 + $V$ \\
conv7\_1 & 4 & .5 & 1 &  256\plusone/64  & 2 & 1 & conv6\_2 + conv1\_2 + $V$ \\
conv7\_2 & 3 &  1 & 1 &   64\plusone/64  & 1 & 1 & conv7\_1 + $V$ \\
conv7\_3 & 1 &  1 & 1 &   64\plusone/67  & 1 & 1 & conv7\_2 + $V$ \\
\bottomrule
\end{tabular}
\end{table}


\begin{table}[bp]
    \centering
    \caption{Hyperparameters for our dataset generation and training.
    }
    \label{tab:parameters}
    \setlength{\tabcolsep}{10pt}%
    \begin{tabular}{ l r }
    	\toprule
    	Parameter                                                                   & Value                   \\ \midrule
    	Training / Test target views                                                & 63,000 / 27,000         \\
    	Learning rate                                                               & 0.0002                  \\
    	Batch size                                                                  & 1                       \\
    	Epochs                                                                      & 4                       \\
    	Optimizer                                                                   & Adam with $\beta=0.9$   \\ \midrule
    	ODS baseline                                                                & 0.064 metres            \\
    	MSI radii                                                                   & [1, 100] metres         \\
    	Target view offset $(x,y,z)$                                                & [0.02, 0.36] metres     \\
    	Temporal rotation $(\theta,\phi,\psi)$                                      & $\pm1.7$\degree         \\
    	Temporal translation $(x,y,z)$                                              & $\pm$0.01 metres        \\
    	$\lambda_{\text{L2}}$, $\lambda_{\text{ERP-L2}}$, or $\lambda_{\text{Per}}$ & 1                       \\
    	$\lambda_{\text{TI}}$                                                       & 10                      \\ \bottomrule
    \end{tabular}
\end{table}


\clearpage
\subsubsection{Rendering Pseudocode}
\label{sec:rendering}

In \cref{alg:render}, we include pseudocode for the $\text{Render}$ function from Section~3.3 in the main paper, which is used to train our network architecture.
Then, in \cref{alg:unityrender}, we provide pseudocode for our real-time MSI renderer implemented in Unity, which additionally uses high-resolution video input and a joint-bilateral filter to improve quality.

\vspace{2em}
\begin{algorithm}[h]
\SetAlgoLined
\SetKwProg{Render}{Render}{}{end}
\Render{} {
	\KwIn{\\
		$\mathcal{M}$: A $w \times h \times N$ MSI.\\
		$P$: A $4 \times 4$ matrix representing a target pose.}
	\vspace{0.25cm}
	\KwOut{\\
		$\hat{I}$: Rendered ERP image from the target pose.}
	\vspace{0.25cm}
  
	\ForEach{$\mathrm{pixel\ location\ } (u, v) \in I$}{
		$\mathbf{r} \gets \text{GetRay}(u, v, P)$ \\
		$\{\mathbf{p}_i\}_{i=1}^{N} \gets \text{GetIntersections}(\mathbf{r}, \mathcal{M})$ \\
		$\{\mathbf{c}_i\}_{i=1}^{N}, \{\alpha_i\}_{i=1}^{N} \gets \text{Sample}(\mathcal{M}, \{\mathbf{p}_i\}_{i=1}^{N})$ \\
		$\hat{I}(u, v) \gets \text{OverComposite}(\{\mathbf{c}_i\}_{i=1}^{N}, \{\alpha_i\}_{i=1}^{N})$
	}
	\KwRet{$\hat{I}$}
}
\caption{Render function from Section 3.3 (main paper)}
\label{alg:render}
\end{algorithm}

\vspace{2em}
\begin{algorithm}[h]
\SetAlgoLined
\SetKwProg{RTRender}{RTRender}{}{end}
\RTRender{} {
	\KwIn{\\
		$I_L$: A high-resolution $w \times h$ left ODS image.\\
		$I_R$: A high-resolution $w \times h$ right ODS image.\\
		$P$: Pose of the VR headset}
	\vspace{0.25cm}
	\KwOut{\\
		$I$: Rendered perspective image from headset pose}
	\vspace{0.25cm}
  
	$I_L', I_R' \gets \text{AntialiasedDownsample}(I_L, I_R)$ \\
	$\mathcal{M} \gets \text{InferMSI}(I_L', I_R')$ \\
	$\mathcal{M}' \gets \text{JointBilateralUpsample}(\mathcal{M}, I_L, I_R)$ \\
	$\mathcal{S} \gets \emptyset$ \\
	\ForEach{$\mathrm{layer} \ l \in \mathcal{M}'$}{
		$\mathcal{S}_l \gets \text{TextureSphere}(\mathcal{M}', I_L, I_R, l)$ \\
	}
	$I \gets \text{RasterizeWithAlpha}(\mathcal{S}, P)$\\
	\KwRet{$I$}
}
\caption{Real-time rendering pipeline in Unity}
\label{alg:unityrender}
\end{algorithm}

\end{document}